\documentclass[nohyperref]{article}

\usepackage{microtype}
\usepackage{graphicx}
\usepackage{subfigure}
\usepackage{booktabs} 

\usepackage{hyperref}

\usepackage[accepted]{icml2022}

\usepackage{amsmath}
\usepackage{amssymb}
\usepackage{mathtools}
\usepackage{amsthm}
\usepackage[super]{nth}
\usepackage{tikz}
\usepackage{soul}
\usepackage{enumitem}
\usepackage{setspace}

\newcommand*\circledB[1]{\tikz[baseline=(char.base)]{
            \node[shape=circle,fill,inner sep=0.2pt] (char) {\textcolor{white}{#1}};}}

\usepackage[capitalize,noabbrev]{cleveref}

\theoremstyle{plain}

\theoremstyle{definition}

\theoremstyle{remark}

\usepackage[textsize=tiny]{todonotes}

\icmltitlerunning{tinySNN: Towards Memory- and Energy-Efficient Spiking Neural Networks}

\begin{document}

\twocolumn[
\icmltitle{tinySNN: Towards Memory- and Energy-Efficient Spiking Neural Networks}



\icmlsetsymbol{equal}{*}

\begin{icmlauthorlist}
\icmlauthor{Rachmad Vidya Wicaksana Putra }{TUW}
\icmlauthor{Muhammad Shafique}{NYUAD}
%
\end{icmlauthorlist}

\icmlaffiliation{TUW}{Department of Computer Engineering, Technische Universit\"at Wien (TU Wien), Vienna, Austria}
\icmlaffiliation{NYUAD}{Division of Engineering, New York University Abu Dhabi (NYUAD), Abu Dhabi, United Arab Emirates}
%

\icmlcorrespondingauthor{Rachmad Vidya Wicaksana Putra}{rachmad.putra@tuwien.ac.at}

\icmlkeywords{Spiking Neural Networks, SNNs, tinySNN, memory efficiency, energy efficiency, machine learning}

\vskip 0.3in
\vspace{-0.2cm}
]



\printAffiliationsAndNotice{}  

\begin{spacing}{1}
\begin{abstract}
Larger Spiking Neural Network (SNN) models are typically favorable as they can offer higher accuracy. 
However, employing such models on the resource- and energy-constrained embedded platforms is inefficient.
Towards this, we present a tinySNN framework that optimizes the memory and energy requirements of SNN processing in both the training and inference phases, while keeping the accuracy high.
It is achieved by reducing the SNN operations, improving the learning quality, quantizing the SNN parameters, and selecting the appropriate SNN model.
Furthermore, our tinySNN quantizes different SNN parameters (i.e., weights and neuron parameters) to maximize the compression while exploring different combinations of quantization schemes, precision levels, and rounding schemes to find the model that provides acceptable accuracy.
The experimental results demonstrate that our tinySNN significantly reduces the memory footprint and the energy consumption of SNNs without accuracy loss as compared to the baseline network. 
Therefore, our tinySNN effectively compresses the given SNN model to achieve high accuracy in a memory- and energy-efficient manner, hence enabling the employment of SNNs for the resource- and energy-constrained embedded applications. 
\end{abstract}
\end{spacing}

\begin{spacing}{1}
\vspace{-0.6cm}
\section{Introduction}
\label{Sec_Intro}

\begin{figure}[t]
\centering
\includegraphics[width=0.87\linewidth]{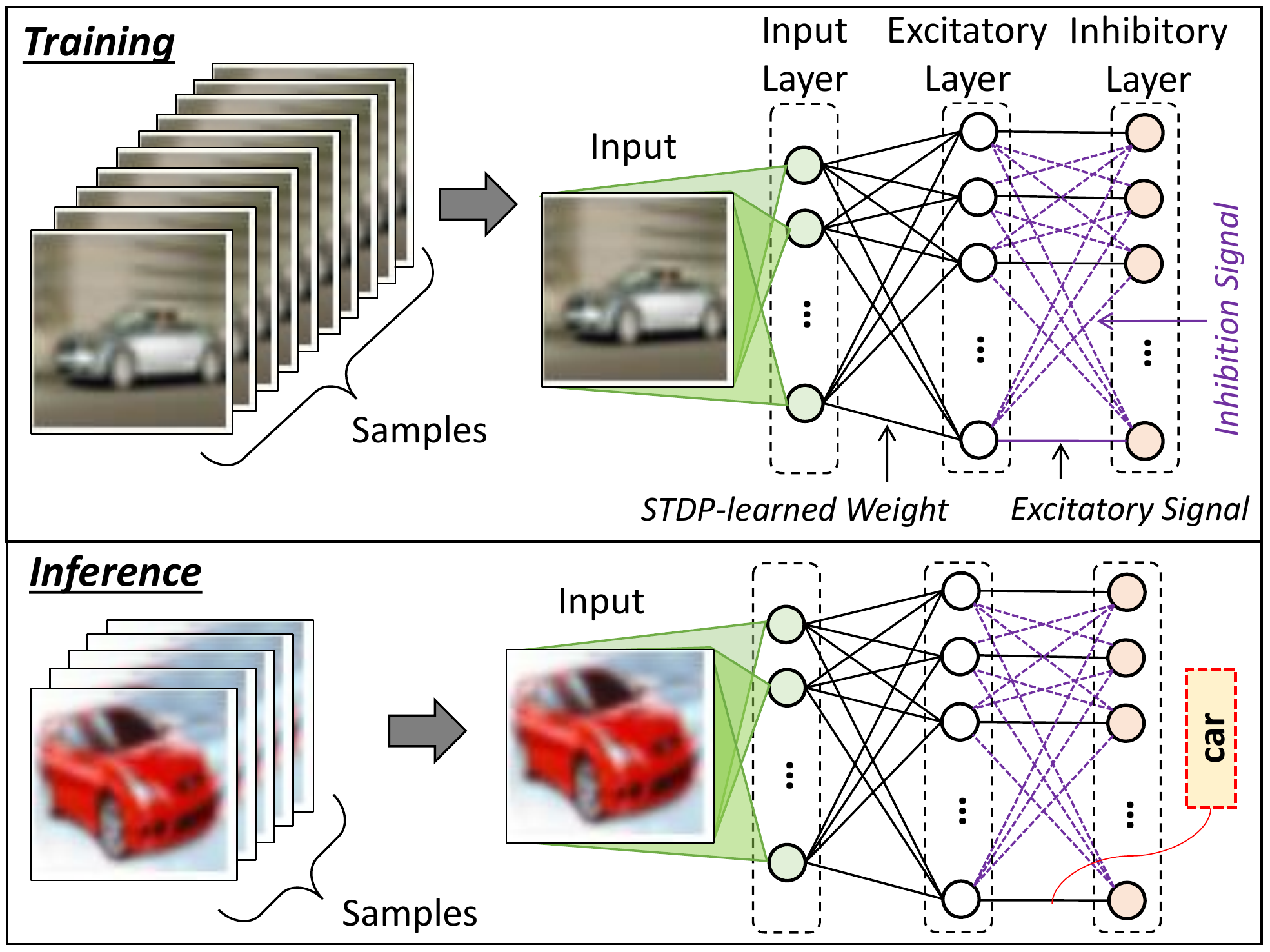}
\vspace{-0.3cm}
\caption{Training and inference phases for SNNs. We consider the fully-connected network as it offers high accuracy under various unsupervised STDP-based learning rules~\cite{Ref_Diehl_STDPmnist_FNCOM15}.}
\label{Fig_SNNs}
\vspace{-0.8cm}
\end{figure}

Spiking Neural Networks (SNNs) have emerged as the neural network models that offer high accuracy under ultra-low power/energy consumption and unsupervised learning settings due to their bio-plausible spike-based operations and learning rule, e.g., Spike-Timing-Dependent Plasticity (STDP)~\cite{Ref_Putra_FSpiNN_TCAD20}.
In recent years, many SNN models have been developed to solve various machine learning tasks, such as object classification and hand gesture recognition~\cite{Ref_Pfeiffer_DLSNN_FNINS18, Ref_Tavanaei_DLSNN_Neunet18,Ref_Diehl_STDPmnist_FNCOM15, Ref_Hazan_SOMSNN_IJCNN18,Ref_Saunders_STDPpatch_IJCNN18,Ref_Saunders_LCSNN_NeuNet19,Ref_Kaiser_DECOLLE_FNINS20,Ref_Putra_ReSpawn_ICCAD21,Ref_Putra_SoftSNN_arXiv22}. 
Recent trends suggest that large SNN models (i.e., a large number of synapses and neurons) are frequently used in state-of-the-art works since they can achieve higher accuracy than the smaller ones due to their capabilities for recognizing more features~\cite{Ref_Putra_SparkXD_DAC21, Ref_Shafique_EdgeAI_ICCAD21}. 
For instance, in a fully-connected SNN  (shown in Fig.~\ref{Fig_SNNs}) with 32-bit floating-point format (FP32) consumes $\sim$200MB of memory and achieves $\sim$92\% accuracy on the MNIST dataset, while a smaller model with $\sim$1MB of memory achieves $\sim$75\% accuracy~\cite{Ref_Putra_FSpiNN_TCAD20}. 
Therefore, the state-of-the-art SNNs usually have a large number of parameters (e.g., weights and neuron parameters) which occupies a large memory footprint. 
This leads to intensive memory accesses which dominate the energy consumption of neural network-based computations~\cite{Ref_Putra_ROMANet_TVLSI21, Ref_Putra_DRMap_DAC20, Ref_Krithivasan_SpikeBundle_ISLPED19}, thereby hindering the deployment of SNNs on the resource- and energy-constrained embedded platforms, e.g., the battery-operated IoT-Edge devices.

\textbf{Targeted Problem:}
\textit{How can we effectively optimize SNNs to reduce their memory and energy requirements, while keeping their accuracy high.} 

\vspace{-0.2cm}
\subsection{State-of-the-art Works and Their Limitations}
\label{Sec_Intro_SOTA}

\begin{figure}[hbtp]
\vspace{-0.2cm}
\centering
\includegraphics[width=\linewidth]{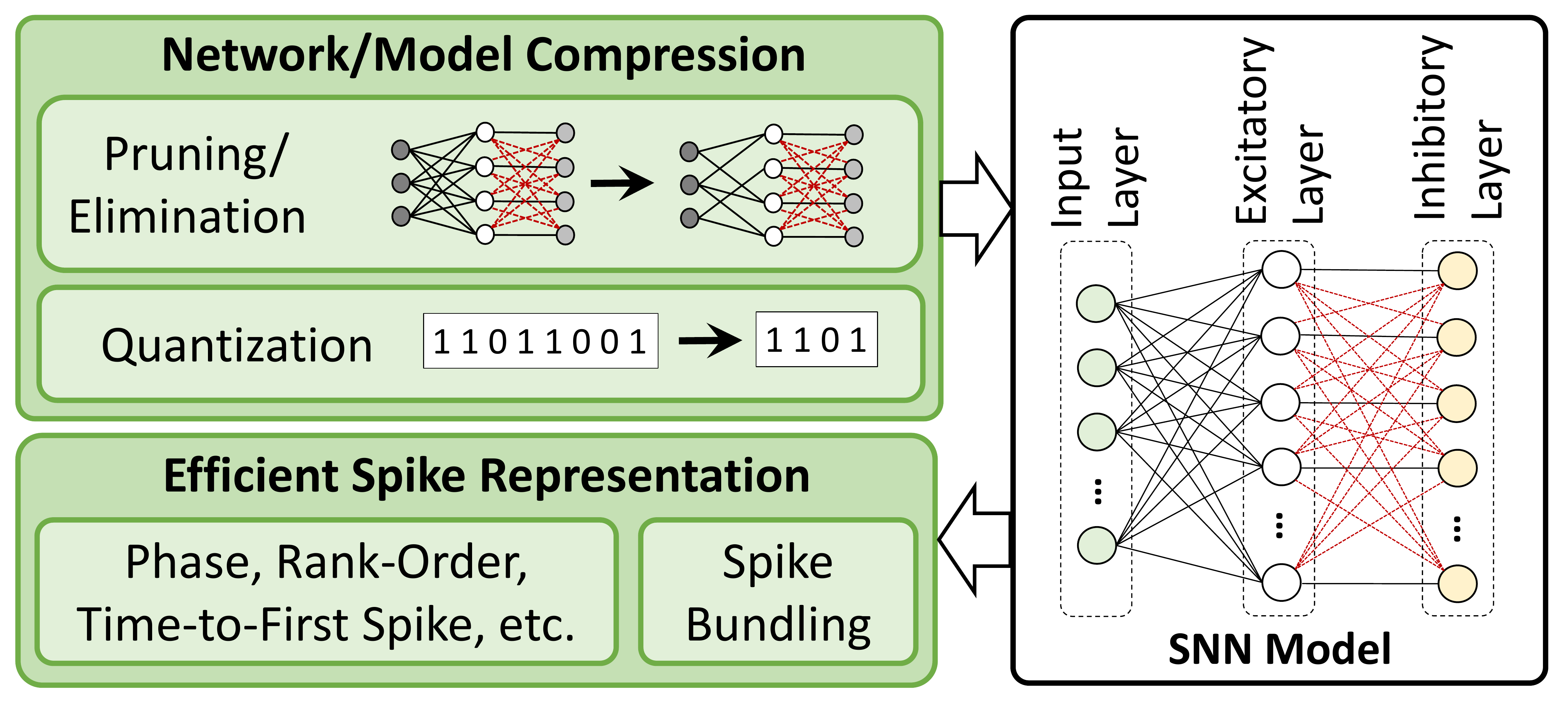}
\vspace{-0.8cm}
\caption{The existing techniques for optimizing the memory footprint of SNNs that leads to the reduction of energy consumption.}
\label{Fig_OverviewTech}
\vspace{-0.2cm}
\end{figure}

Fig.~\ref{Fig_OverviewTech} highlights the state-of-the-art techniques for optimizing the memory footprint of SNNs that lead to energy saving. 
Works of~\cite{Ref_Rathi_PruneQuantizeSNN_TCAD18, Ref_Hu_QuantSTDPSNN_NeCo21} perform quantization and/or weight pruning to compress the SNN model for inference. 
However, it still requires a huge number of (inhibitory) neurons to effectively perform inference, and suffers from accuracy degradation as compared to the baseline model (i.e., dense and non-quantized model).  
Other works employ different spike coding schemes to optimize the memory footprint for storing the information of spikes~\cite{Ref_Kayser_PhaseCoding_Neuron09, Ref_Park_T2FSNN_DAC20}. 
The work of~\cite{Ref_Krithivasan_SpikeBundle_ISLPED19} reduces the spike memory accesses by bundling a sequence of spikes into a single spike. 
However, these spike representation techniques do not optimize the SNN parameters, which dominate the memory footprint and access energy.
Other works mainly focus on improving the accuracy, but at the cost of additional computations which leads to high memory and energy requirements~\cite{Ref_Srinivasan_EnhPlast_IJCNN17,Ref_Hazan_SOMSNN_IJCNN18,Ref_Panda_ASP_JETCAS18,Ref_Saunders_LCSNN_NeuNet19,Ref_Hazan_LMSNN_AMAI19}. 
For instance, the work of~\cite{Ref_Srinivasan_EnhPlast_IJCNN17} improves the STDP-based learning by ensuring that the update is essential. 
However, it also requires a huge number of (inhibitory) neurons to effectively perform inference.
Therefore, the state-of-the-art techniques offer partial benefits (i.e., either memory reduction, energy saving, or accuracy improvement) thereby making them unsuitable for embedded SNN applications which necessitate all the above benefits.
Moreover, these techniques are not suitable for systems that require an efficient online training, which requires a lightweight training mechanism at run time~\cite{Ref_Putra_SpikeDyn_DAC21,Ref_Putra_lpSpikeCon_arXiv22}.

In summary, there is a significant need for \textit{an optimization framework that reduces the memory and energy requirements of SNNs for both the training and inference phases while avoiding accuracy loss}. 

\vspace{-0.1cm}
\subsection{Contributions of This Work}
\label{Sec_Intro_Contrib}
\vspace{-0.1cm}

To overcome the above challenges, we propose \textit{\textbf{tinySNN}, a framework that optimizes the memory and energy requirements of SNNs for training and inference while maintaining accuracy}.
Our tinySNN employs the following steps.
\vspace{-0.1cm}
\begin{itemize}[leftmargin=*]
    \vspace{-0.2cm}
    \item \textbf{Reducing the SNN operations (Section~\ref{Sec_TinySNN_OpsReduction}):}
    It replaces the inhibitory neurons with the direct lateral inhibition. 
    Hence, the inhibitory neurons' parameters and operations, as well as the respective connections are eliminated.
    \vspace{-0.2cm}
    \item \textbf{Enhancing the learning quality (Section~\ref{Sec_TinySNN_Learning}):} 
    It improves the learning rule to compensate the reduction of SNN operations 
    by minimizing the spurious weight updates and employing adaptive potentiation in training. 
    \vspace{-0.2cm}
    \item \textbf{Quantizing the SNN parameters (Section~\ref{Sec_TinySNN_Quantize}):}
    It quantizes different parameters (i.e., weights and neuron parameters) and explores different combinations of quantization schemes, precision levels, and rounding schemes to find the configuration that meets the design constraints.
    \vspace{-0.2cm}
    \item \textbf{SNN model selection (Section~\ref{Sec_TinySNN_ModelSelect}):}
    It quantifies the memory-accuracy trade-off from the given SNN models using the proposed reward function, and then chooses the one with the highest score.
    \vspace{-0.2cm}
\end{itemize}

\section{tinySNN Framework}
\label{Sec_TinySNN}

\begin{figure}[hbtp]
\centering
\includegraphics[width=\linewidth]{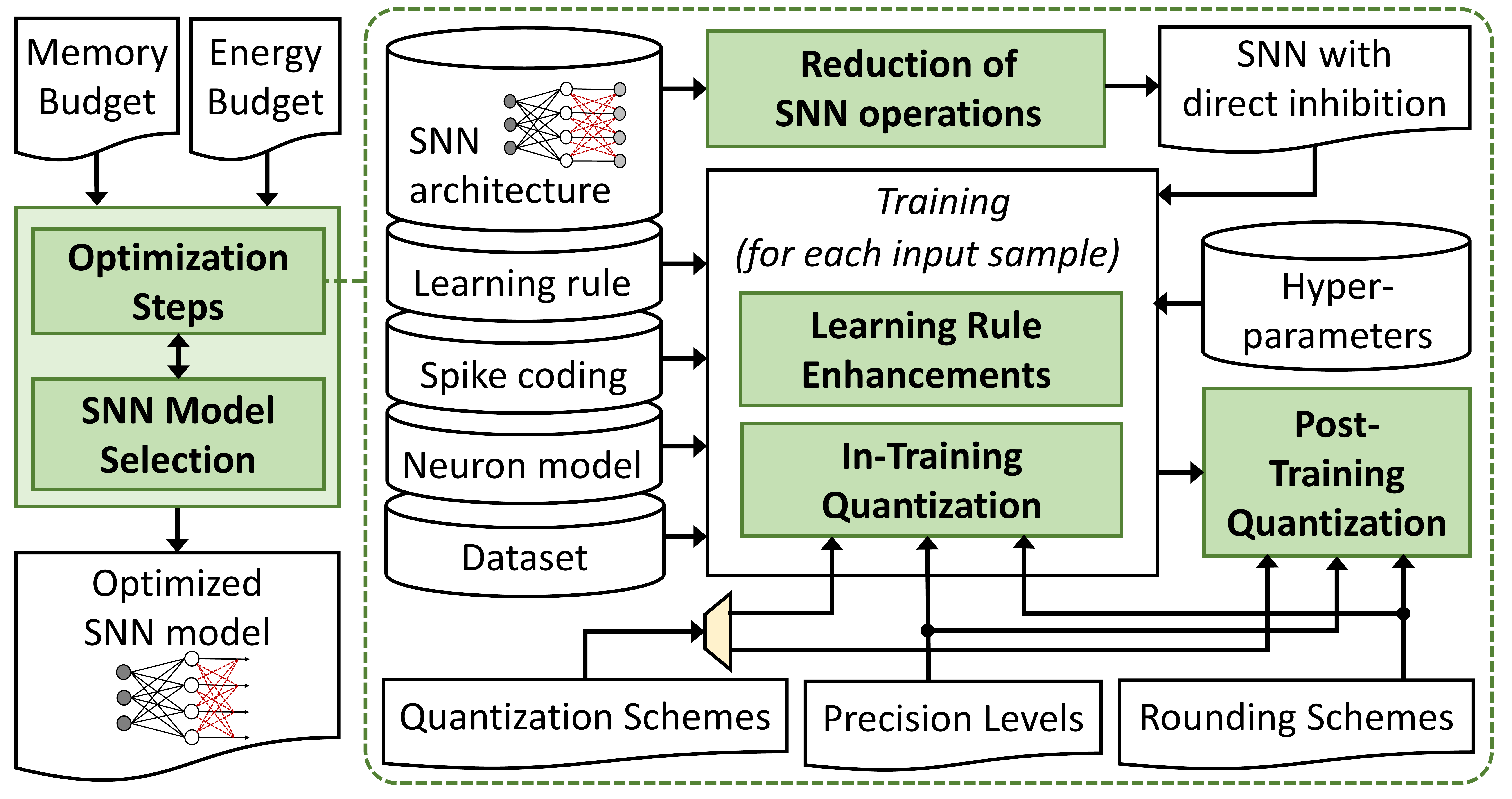}
\vspace{-0.7cm}
\caption{The tinySNN framework showing its key steps. 
}
\label{Fig_OurMethod}
\vspace{-0.3cm}
\end{figure}

Our tinySNN employs several key steps as shown in Fig.~\ref{Fig_OurMethod}. 

\vspace{-0.2cm}

\subsection{Reduction of SNN Operations}
\label{Sec_TinySNN_OpsReduction}

\begin{figure}[hbtp]
\centering
\includegraphics[width=\linewidth]{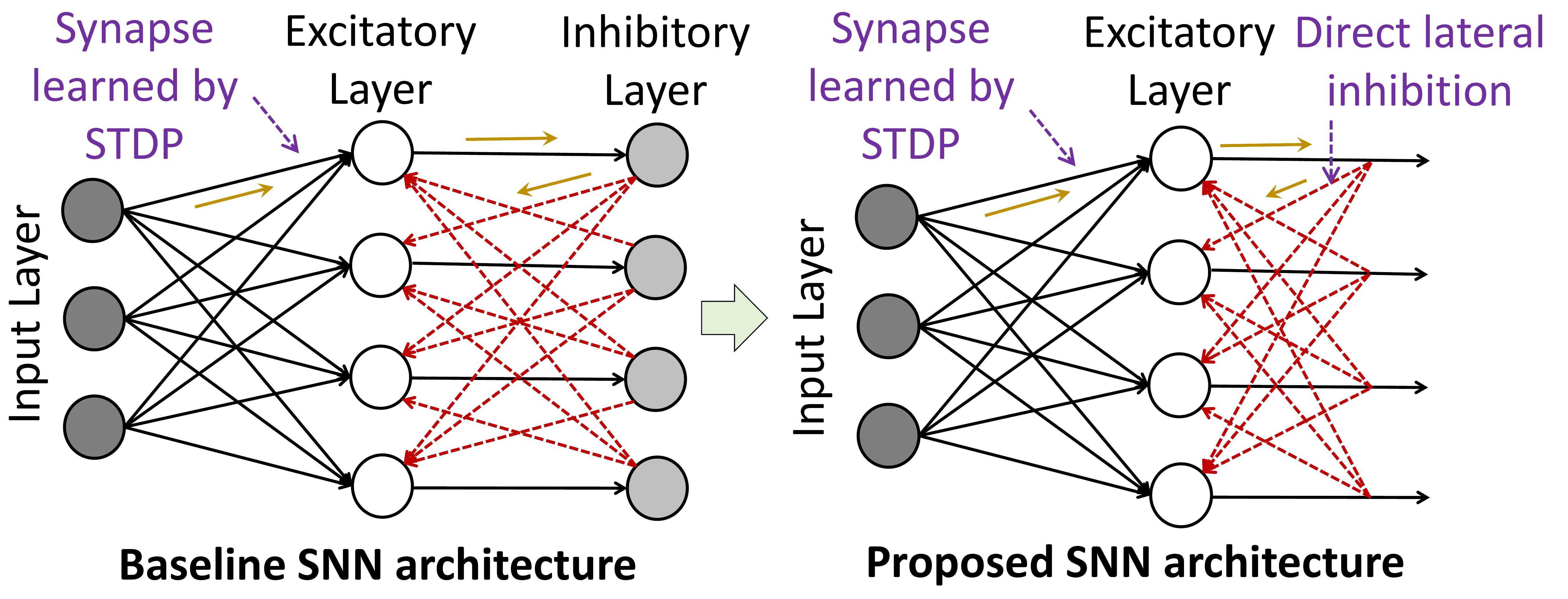}
\vspace{-0.7cm}
\caption{The proposed reduction of SNN operations; adapted from~\cite{Ref_Putra_FSpiNN_TCAD20}.}
\label{Fig_ModifiedSNN}
\vspace{-0.7cm}
\end{figure}

We observe that the inhibitory neurons require different parameter values from the excitatory ones to properly generate inhibitory spikes for effective SNN processing. 
Therefore, the parameters of inhibitory neurons also need to be stored in memory.
This indicates that a large number of inhibitory neurons leads to a large memory footprint, and thereby energy consumption. 
Towards this, \textit{we propose to replace the inhibitory neurons with direct lateral inhibition to decrease the number of neuron operations and parameters, hence curtailing the memory footprint and energy consumption}~\cite{Ref_Putra_FSpiNN_TCAD20}, as shown in Fig.~\ref{Fig_ModifiedSNN}. 
Therefore, the function of inhibitory spikes (i.e., providing competition among neurons) is replaced by spikes that come directly from the excitatory neurons.

\subsection{Learning Rule Enhancements}
\label{Sec_TinySNN_Learning}

We observe that at least one excitatory neuron has to recognize a specific class and produces the highest number of spikes (for the rate coding) to represent its recognition.  
Hence, the information of generated spikes should be considered for improving the accuracy. 
Towards this, \textit{we propose the STDP-based learning enhancements through timestep-based weight updates and adaptive potentiation factor ($k$) that leverage the information of postsynaptic spikes}~\cite{Ref_Putra_FSpiNN_TCAD20}. 
Our weight update happens once within each defined timestep to avoid the spurious updates, while leveraging the number of postsynaptic spikes to determine how strong $k$ should be performed. 
We use the maximum number of postsynaptic spikes ($maxN$) observed during the presentation of input spikes until the update time, then normalize it with the defined threshold ($N_{th}$) to calculate $k$, as shown by Eq.~\ref{Eq_STDP_k}. 
Afterwards, the weight change ($\Delta w$) is calculated using Eq.~\ref{Eq_ImprovedSTDP}, with $\eta_{post}$ denotes the learning rate, $x_{pre}$ denotes the presynaptic trace, $w_m$ denotes the maximum weight value, and $w$ denotes the previous weight value.
In this manner, the confidence level of learning increases over time during the presentation of each input image.

\vspace{-0.2cm}
\begin{equation}
\vspace{-0.2cm}
\small
k = \left \lceil \frac{maxN}{N_{th}} \right \rceil\\
\label{Eq_STDP_k}
\end{equation}
\begin{equation}
\small
\Delta w = 
k \eta_{post} x_{pre} (w_{m}-w) \; \; \text{on} \; \text{update time} \\
\label{Eq_ImprovedSTDP}
\end{equation}

\subsection{SNN Parameter Quantization}
\label{Sec_TinySNN_Quantize}

\begin{figure}[t]
\centering
\includegraphics[width=\linewidth]{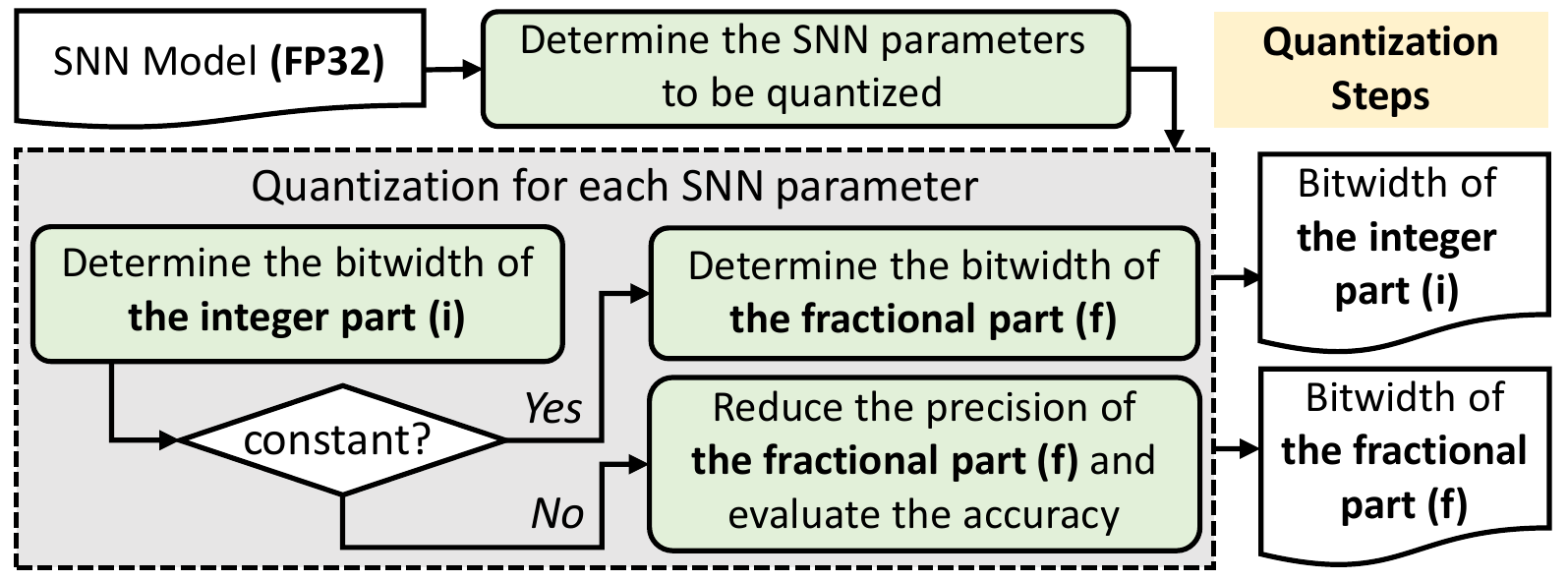}
\vspace{-0.7cm}
\caption{The proposed fix-point quantization steps for a given SNN model; adapted from~\cite{Ref_Putra_QSpiNN_IJCNN21}.}
\label{Fig_Quantize}
\vspace{-0.6cm}
\end{figure}

An SNN model typically has different types of parameters (e.g., weights and neuron parameters) which need to be stored in the memory during the training and inference phases.
We observe that these parameters can be quantized to minimize the memory footprint and the energy consumption.
Towards this, \textit{we propose to quantize different SNN parameters while exploring different combinations of quantization schemes, precision levels, and rounding schemes to find the appropriate model}; see Fig.~\ref{Fig_Quantize}.
For quantization schemes, we consider the post-training quantization (PTQ) and the quantization-aware training/in-training quantization (ITQ)~\cite{Ref_Krishnamoorthi_Whitepaper_arXiv18}. 
For rounding schemes, we consider the truncation, round-to-the-nearest, and stochastic rounding~\cite{Ref_Hopkins_Rounding_RSTA20}.
The idea of our quantization is to maximize the compression for each parameter, and define the precision level based on its impact to the accuracy. 
To do this, we first select the parameters to be quantized for the given model (FP32). 
Then, we analyze the significance of each parameter for defining the integer and fractional bitwidth. 
\textit{For integer part}, we observe the range of values for the parameter during SNN processing, then determine the the bitwidth requirement accordingly.
\textit{For fractional part}, if the parameter is a constant, the bitwidth follows the parameter value; while if the parameter is a variable, the bitwidth is analyzed by reducing the precision and observing the accuracy. 
In this work, for our case study, we quantize weights and two neuron parameters, i.e., membrane potential ($V_{mem}$) and threshold potential ($V_{th}$).

\vspace{-0.2cm}
\subsection{SNN Model Selection}
\label{Sec_TinySNN_ModelSelect}

Once we obtain the compressed models from different combinations of optimization techniques, we need to select the appropriate model that provides high accuracy and meets the memory and energy constraints. 
Towards this, \textit{we propose a model selection technique that quantifies the trade-off between accuracy and memory of the given model using our reward function}, which can be expressed as Eq.~\ref{Eq_Reward}.
Here, $acc_q$ is the accuracy of the quantized model, $M_{norm}$ is the normalized memory, and $\mu$ is the trade-off coefficient whose value is the non-negative real number.
$M_{norm}$ is the ratio between the size of quantized model ($M_q$) and non-quantized model ($M_0$).
Furthermore, the memory footprint ($M$) is the total size of the weights ($Mw$) and neuron parameters ($Mn$).  
$Mw$ equals to the number of weights ($Nw$) multiplied by the corresponding bitwidth ($Bw$).
Meanwhile, $Mn$ equals to the number of neuron parameter-\textit{k} ($Nn_k$) multiplied by the corresponding bitwidth ($Bn_k$).
In addition, the energy consumption characteristics are also considered to select the SNN model that meets the memory- and energy-constraints. 
\vspace{-0.2cm}
\begin{equation}
\vspace{-0.2cm}
\small
R = acc_q - \mu \cdot M_{norm} \;\;\;\; \text{with} \;\;\;\; M_{norm} = \frac{M_q}{M_0}  
\label{Eq_Reward}
\end{equation}
\begin{equation}
\small
\begin{split}
M = Mw + Mn = Nw \cdot Bw + \sum_k Nn_k \cdot Bn_k
\end{split}
\label{Eq_Memory}
\vspace{-0.2cm}
\end{equation}

\vspace{-0.2cm}
\section{Evaluation Methodology}
\label{Sec_EvalMethod}
\vspace{-0.1cm}

We use a PyTorch-based simulation~\cite{Ref_Hazan_BindsNET_FNINF18} for evaluating the accuracy of SNN models, which runs on GPUs (i.e., Nvidia GeForce RTX 2080 Ti~\cite{Ref_RTX2080Ti} for GPGPU and Nvidia Jetson Nano~\cite{Ref_JetsonNano} for embedded GPU).
We consider the fully-connected SNN with 400 excitatory neurons and the rate coding. 
For workloads, we use the MNIST and Fashion MNIST datasets~\cite{Ref_Lecun_MNIST_IEEE98,Ref_Xiao_FMNIST_arXiv17}. 
For comparisons, we recreate the SNN design with the pair-wise STDP (as the baseline)~\cite{Ref_Diehl_STDPmnist_FNCOM15} and the SNN design with Self-Learning STDP (SL-STDP)~\cite{Ref_Srinivasan_EnhPlast_IJCNN17}. 
To evaluate the memory, we leverage the weights, the neuron parameters, and the respective bitwidth.
To evaluate the energy consumption, we leverage the simulation time and the operational power.

\section{Results and Discussion}
\label{Sec_Results}

\subsection{Maintaining the Accuracy}
\label{Sec_Results_Accuracy}
\vspace{-0.1cm}

\begin{figure}[hbtp]
\vspace{-0.2cm}
\centering
\includegraphics[width=\linewidth]{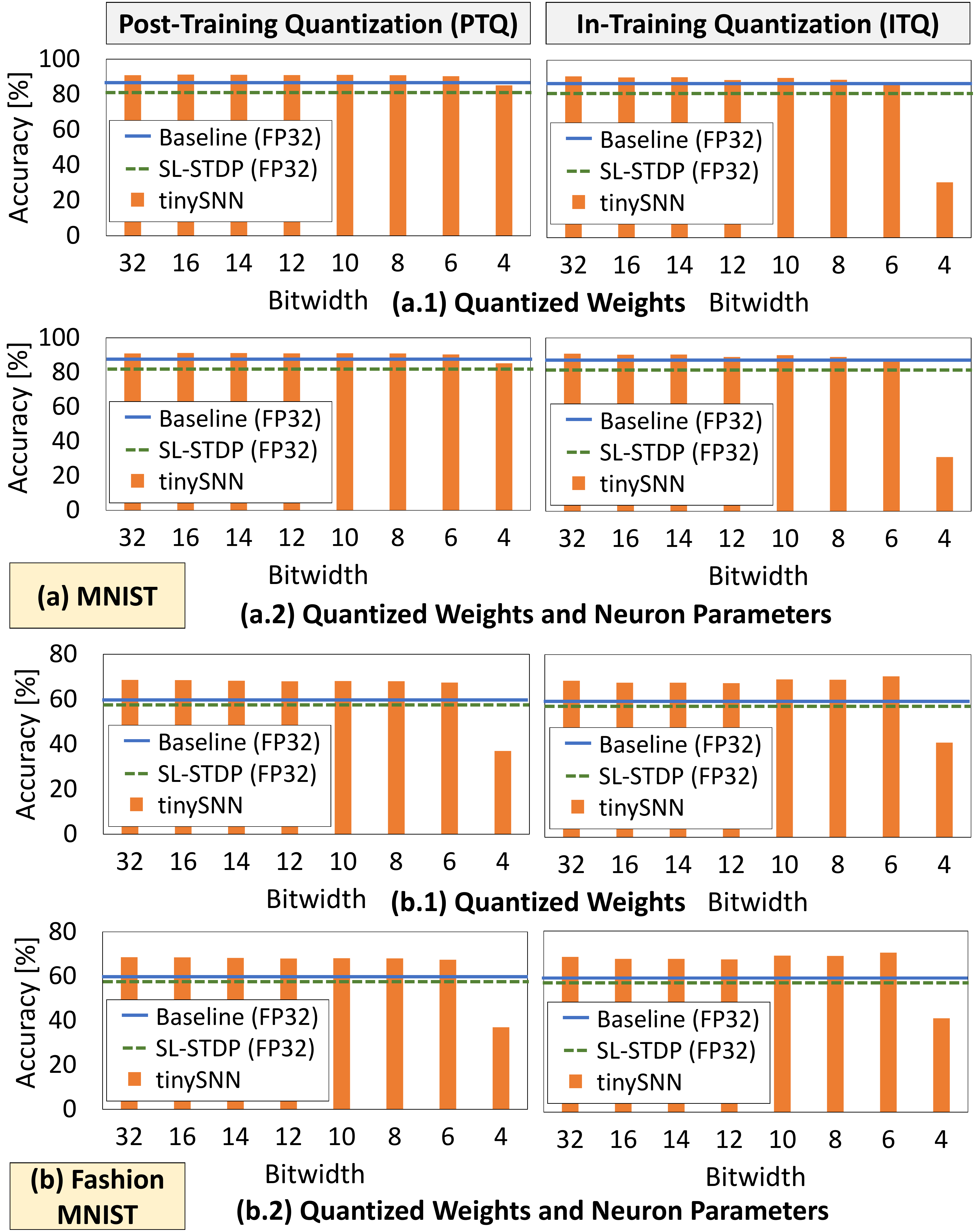}
\vspace{-0.7cm}
\caption{Accuracy of the baseline SNN (FP32), the SL-STDP (FP32), and our tinySNN for (a) the MNIST and (b) the Fashion MNIST. 
Here, our tinySNN employs the truncation rounding.}
\label{Fig_Results_Accuracy}
\vspace{-0.3cm}
\end{figure}

Fig.~\ref{Fig_Results_Accuracy} presents experimental results for accuracy.
These results show that our tinySNN with precision level $\geq$ 8 bits improves the accuracy over the baseline (FP32) and the SL-STDP (FP32) across different quantization schemes (i.e., PTQ and ITQ) and datasets.
The accuracy improvement can be associated with several reasons. 
First, the learning enhancements in our tinySNN effectively perform weight updates by avoiding spurious updates and employing adaptive learning rate based on the observation on the postsynaptic spikes.
Second, the 8-bit precision (or more) in our tinySNN provides a sufficient range of weight values that induces each neuron to recognize a specific class.  

\vspace{-0.1cm}
\subsection{Reducing the Memory Footprint}
\label{Sec_Results_Memory}

\begin{figure}[hbtp]
\vspace{-0.2cm}
\centering
\includegraphics[width=0.8\linewidth]{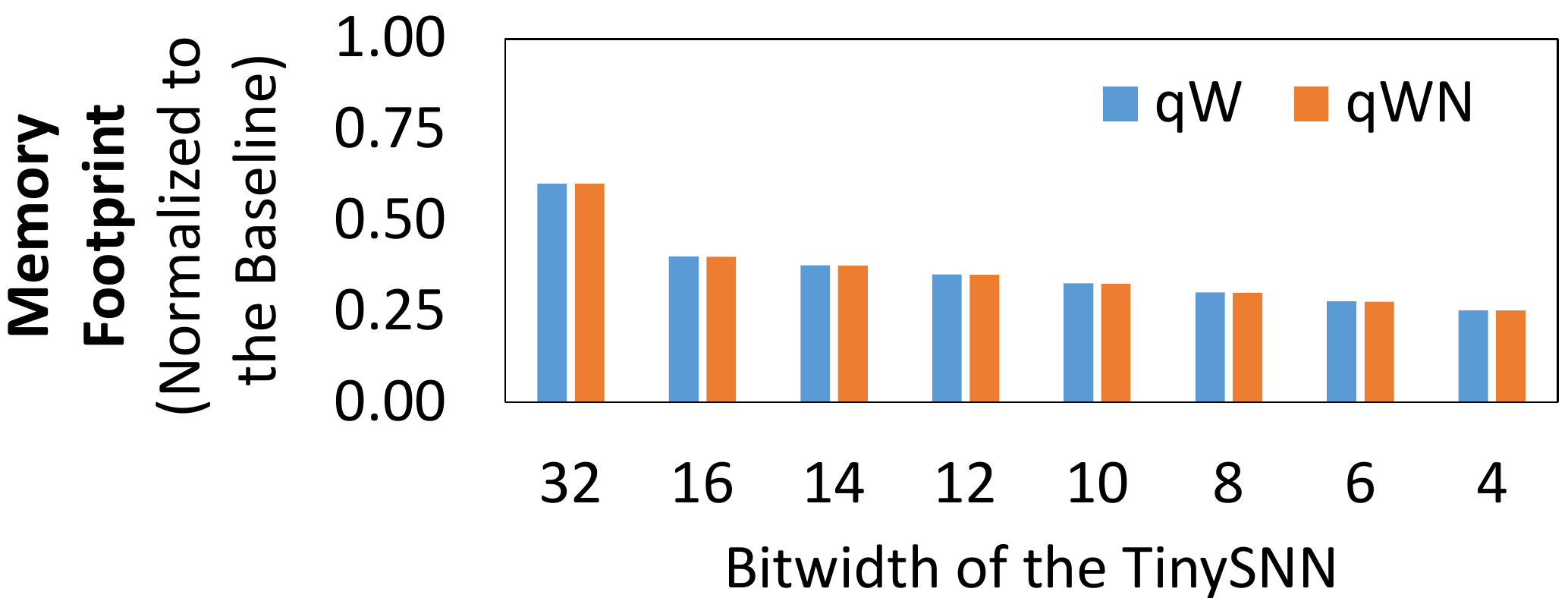}
\vspace{-0.2cm}
\caption{Memory footprint of our tinySNN, across different precision levels and quantized parameters, i.e., quantized weights (qW), quantized weights and neuron parameters (qWN). 
}
\label{Fig_Results_Memory}
\vspace{-0.4cm}
\end{figure}

Fig.~\ref{Fig_Results_Memory} presents the experimental results for memory footprint.
These results show that our tinySNN effectively reduce the size of SNN models, i.e., up to 3.3x and 3.32x memory savings for 8-bit weights (qW) and for 8-bit weights and neuron parameters (qWN), respectively. 
These savings mainly come from the elimination of the inhibitory neurons and the reduction of weight precision, while additional savings come from the quantization of neuron parameters.

\subsection{Reducing the Energy Consumption}
\label{Sec_Results_Energy}

\begin{figure}[hbtp]
\vspace{-0.2cm}
\centering
\includegraphics[width=\linewidth]{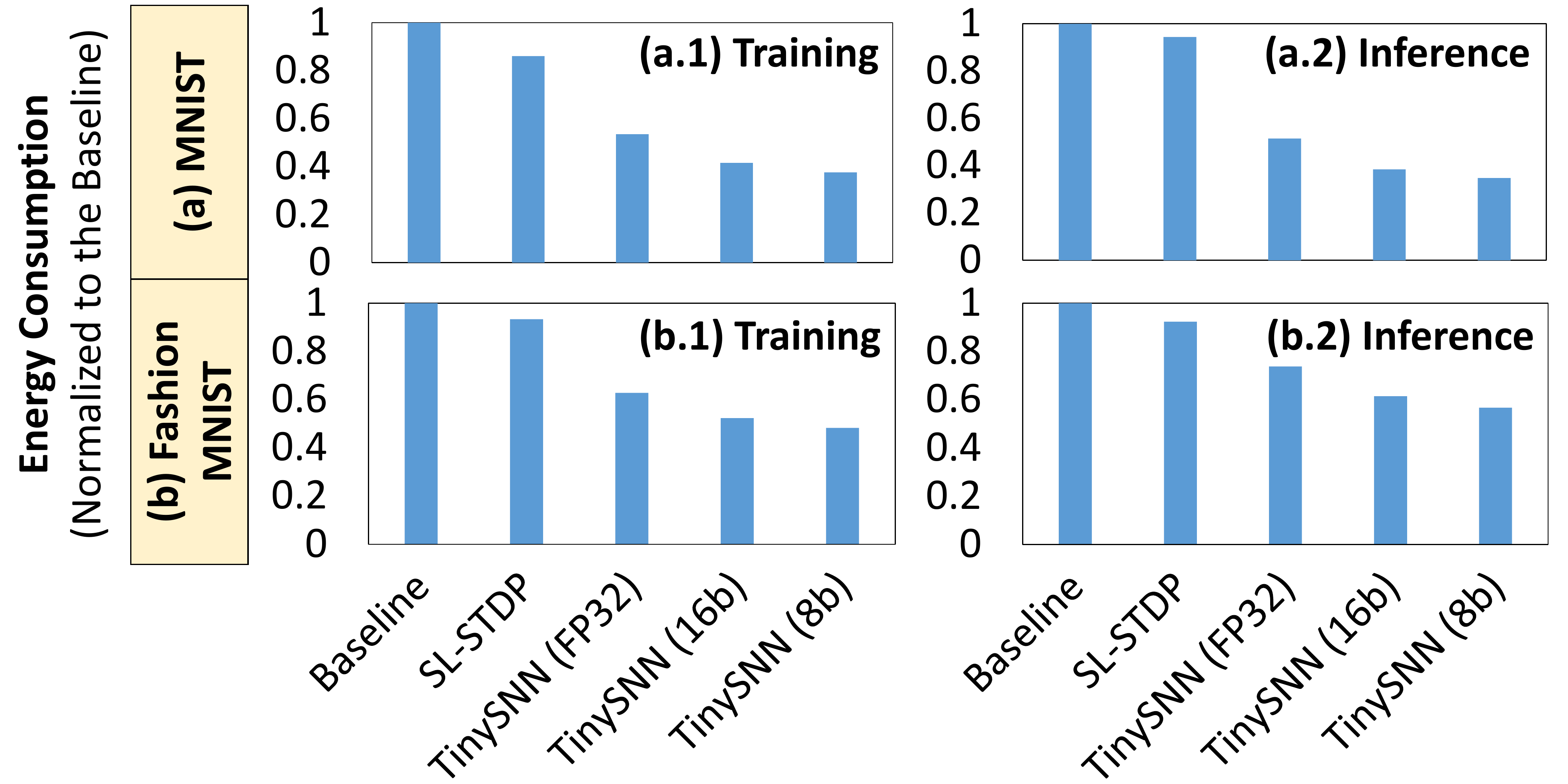}
\vspace{-0.8cm}
\caption{Energy consumption of SNN processing in training and inference for (a) the MNIST and (b) the Fashion MNIST.}
\label{Fig_Results_Energy}
\vspace{-0.4cm}
\end{figure}

Fig.~\ref{Fig_Results_Energy} presents the experimental results for energy consumption of SNN processing.
These results show that our tinySNN effectively reduces the energy consumption of SNNs for both the training and inference phases.
In the training phase, our tinySNN (FP32) optimizes the energy consumption by up to 1.9x since it eliminates the inhibitory neuron operations and simplify the learning rule.
Then, the quantization optimizes the energy consumption even further, e.g., by up to 2.7x for tinySNN (8-bit).   
In the inference phase, our tinySNN (FP32) optimizes the energy consumption by up to 1.9x mainly due to the elimination of the inhibitory neuron operations.
Then, the quantization decreases the energy consumption even more, e.g., by up to 2.9x for tinySNN (8-bit).    

\subsection{Selecting the SNN Model}
\label{Sec_Results_ModelSelect}

\begin{figure}[hbtp]
\vspace{-0.2cm}
\centering
\includegraphics[width=\linewidth]{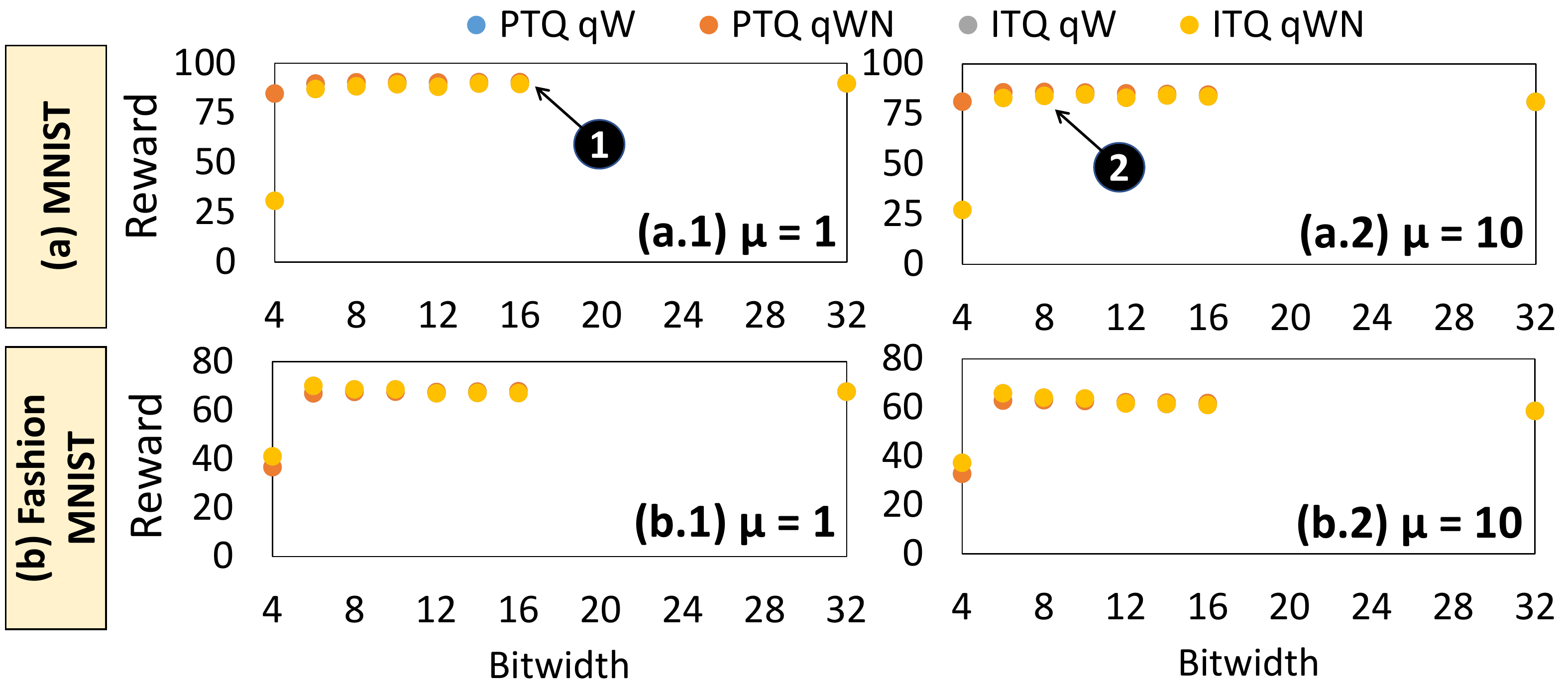}
\vspace{-0.7cm}
\caption{Our SNN model selection explores different quantization schemes (PTQ, ITQ) and quantized parameters (qW, qWN). 
}
\label{Fig_Results_Selection}
\vspace{-0.4cm}
\end{figure}

Fig.~\ref{Fig_Results_Selection} presents the experimental results of SNN model selection in our tinySNN.
These results show that different $\mu$ values result in the different reward values that lead to different selected models.
For instance, a small value of $\mu$ indicates that the trade-off gives a higher priority to the accuracy than the memory.
In such a case, a large SNN model (with a high precision level) will be selected by the algorithm, as shown by~\circledB{1}.
Meanwhile, a larger value of $\mu$ indicates that the trade-off gives a higher priority to the memory than the accuracy. 
In such a case, a smaller SNN model (with a lower precision level) will be selected by the algorithm, as shown by~\circledB{2}.
In addition, the energy consumption characteristics from Section~\ref{Sec_Results_Energy} are also considered to judiciously select the SNN model that achieves the expected accuracy, while meeting the memory- and energy-constraints.

\section{Conclusion}
\label{Sec_Conclusion}

In this paper, we present a tinySNN framework to obtain the memory- and energy-efficient SNNs in both the training and inference phases, while keeping the accuracy high.
Furthermore, our tinySNN also finds the appropriate model for the given memory- and energy-constraints.
In this manner, our tinySNN may enable the applicability of SNNs on diverse embedded applications.






\bibliography{references}
\bibliographystyle{icml2022}


\end{spacing}

\end{document}